\documentclass[letterpaper, 10pt, journal]{IEEEtran}
\IEEEoverridecommandlockouts  
\usepackage[utf8]{inputenc}
\usepackage[english]{babel}
\usepackage{lscape}
\usepackage{pifont}
\usepackage{longtable}
\usepackage{booktabs}
\usepackage{graphicx}
\usepackage{multirow}
\usepackage{setspace} 
\usepackage{hyperref}
\usepackage{amsmath,amssymb,amsfonts}
\usepackage{cite}
\usepackage{color}

\usepackage[margin=1in]{geometry}
\usepackage[caption=false, font=footnotesize]{subfig}
\hypersetup{
    colorlinks=true,
    linkcolor=blue,
    filecolor=magenta,      
    urlcolor=blue,
    citecolor=magenta
}
\definecolor{gold}{rgb}{0.9290, 0.6940, 0.1250}
\definecolor{redd}{rgb}{0.8500, 0.3250, 0.0980}
\definecolor{grey}{rgb}{0.6, 0.6, 0.6}
\newtheorem{thm}{Theorem}

\newtheorem{asm}{Assumption}
\newtheorem{defn}{Definition}

\newtheorem{rem}{Remark}

\title{An Event-Triggered Framework for Trust-Mediated Human-Autonomy Interaction}
\author{Daniel A. Williams$^{1}$, Airlie Chapman$^{2}$, Chris Manzie$^{1}$ 
\thanks{$^{1}$Daniel A. Williams and Chris Manzie are with the Department of Electrical Engineering, The University of Melbourne, Parkville, Australia; Williams is supported by an Australian Government Research Training Program Scholarship.}
\thanks{$^{2}$Airlie Chapman is with the Department of Mechanical Engineering,
        The University of Melbourne, Parkville, Australia.}
}
\date{}
\begin{document}
\maketitle
\thispagestyle{empty}
\pagestyle{empty}

\begin{abstract}
Inspired by the increased cooperation between humans and autonomous systems, we present a new hybrid systems framework capturing the interconnected dynamics underlying these interactions.
The framework accommodates models arising from both the autonomous systems and cognitive psychology literature in order to represent key elements such as human trust in the autonomous system.
The intermittent nature of human interactions are incorporated by asynchronous event-triggered sampling at the framework's human-autonomous system interfaces.
We illustrate important considerations for tuning framework parameters by investigating a practical application to an autonomous robotic swarm search and rescue scenario.
In this way, we demonstrate how the proposed framework may assist in designing more efficient and effective interactions between humans and autonomous systems.
\end{abstract}

\section{Introduction}

\begin{figure*}[ht]
    \centering
    \includegraphics[width=2\columnwidth]{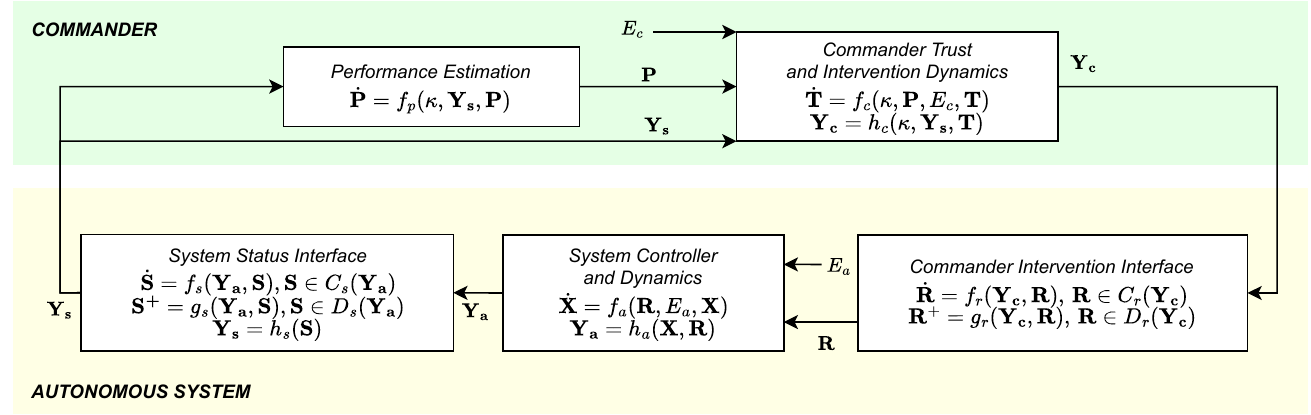}
    \caption{Overview of the proposed HAI framework.}
    \label{fig:framework}
\end{figure*}

With autonomous systems increasingly appearing in commercial and humanitarian applications, there is a need for a paradigm to model the complete human-autonomous system interactions (`HAI').
Key aspects of HAI include intermittent status updates \cite{wang_co-design_2017}, task performance estimation \cite{nam_predicting_2017}, human trust dynamics \cite{nam_predicting_2017,lee_trust_1992,schaefer_measuring_2016}, and trust-driven interventions \cite{xu_optimo_2015,akash_dynamic_2017,liu_clustering_2021}.
While there is currently no standard closed-loop model containing the requisite dynamics at sufficient generality to accommodate most models of HAI dynamics, systems theoretic frameworks can help to derive mathematical guarantees about the closed-loop operation of the entire system without prescribing the individual subsystems exactly \cite{nesic_framework_2013,poveda_framework_2017,garcia-rosas_personalized_2021}.
For HAI in particular, these guarantees can be used to complement existing insights from cognitive psychology into the evolution of trust between the human and autonomous system during a given interaction.

A notion of trust arises in HAI because of the autonomous system's role in acting on behalf of and providing information to a supervising human (whom we term the `commander').
As an illustrative example we consider a commander supervising a team of autonomous robots searching for survivors in a disaster zone, with the commander able to adjust the team's formation in response to the team's performance and the terrain.
With sufficient trust, a commander feels more confident about personal safety and the system's capabilities \cite{schaefer_measuring_2016}, allowing better delegation of task responsibility to the autonomous system \cite{jian_foundations_2010,nahavandi_trusted_2017}.
A considerable body of literature has consequently emerged around trust in HAI \cite{shahrdar_survey_2019}, in particular regarding autonomous swarms \cite{drew_multi-agent_2021}, system transparency \cite{hepworth_human-swarm-teaming_2021} and autonomy levels \cite{nam_models_2020}.
An open challenge lies in integrating representative models of trust and intervention with autonomous system dynamics into a formal systems theoretic framework in order to analyze the overall closed-loop operation.

Another crucial aspect of closed-loop framework operation is the flow of information, specifically the choice of a sampling scheme for communication between the commander and autonomous system.
Periodic sampling relies on a fixed sample rate that is high enough to minimize sampling errors, but which can incur transmission costs through redundant samples during periods of steady output \cite{tabuada_event-triggered_2007}. 
Event-triggered sampling (ETS) can balance sampling error and transmission frequency by triggering samples only when the sampling error exceeds a threshold \cite{astrom_event_2008}. 
Intuitively this approach allows humans to monitor and control autonomous systems intermittently as opposed to periodically, allowing the reallocation of cognitive resources to secondary tasks (e.g. concurrent with monitoring the team's search, the commander may need to communicate with other humans and coordinate the rescue of identified survivors).

In \cite{tabuada_event-triggered_2007} a state-dependent error threshold is introduced for triggering samples of continuous-time control systems, and the application of such schemes to linear and nonlinear control systems is analyzed in detail. 
Building on this, in \cite{postoyan_framework_2015} a hybrid system formulation of ETS is devised for a non-linear plant and controller that includes auxiliary variables for sampler triggering.
Sufficient conditions are then derived for guaranteeing the stability of the origin for different applications of the ETS framework, however there is an implicit requirement for simultaneous updates of both samplers. 
For scenarios in which the plant and controller subsystems communicate intermittently, asynchronous ETS schemes which do not require simultaneous state samples \cite{mazo_asynchronous_2014} are more appropriate.
Given a need for asynchronous ETS of non-linear systems with uniform global asymptotic stability \cite{wang_unifying_2020}, no Zeno behavior between triggers \cite{abdelrahim_robust_2017} and allowing sampling methods other than the zero-order hold \cite{wang_unifying_2020}, a framework with these attributes for sampling rates within specified bounds appears in \cite{williams_asynchronous_2024}.

The contributions of this paper are threefold.
First, we propose a general systems theoretical framework for HAI modeling, extending \cite{poveda_framework_2017} by introducing two interfaces for measurement and control of the autonomous system.
We demonstrate how the framework accommodates existing models from control theory, human-automation interaction, and social psychology, enabling systematic reasoning about closed-loop trust dynamics in HAI scenarios. 
Second, we establish uniform global asymptotic stability of solution sets when using a more general form of asynchronous event-triggered sampling than \cite{williams_asynchronous_2024}.
Finally, we implement the proposed scheme for a practical application (swarm search and rescue) and investigate the effect of tuning framework parameters.

\section{HSI Trust Framework}\label{sec:HSITF}

We first propose a framework for modeling interactions between a human commander and an autonomous system, depicted in Figure \ref{fig:framework} as five interconnected dynamic subsystems. Each isolated subsystem is described in the following subsections.

\subsection{Commander Intervention Interface}

We  define a commander's intervention signal $\mathbf{Y_c}\in C_c\subset \mathbb{R}^{c}$.
To accommodate jumps caused by commander interventions, we adopt a hybrid system representation from \cite{goebel_hybrid_2012}.
The commander intervention interface extracts from $\mathbf{Y_c}$ a reference signal $\mathbf{R}\in C_r(\mathbf{Y_c})\cup D_r(\mathbf{Y_c})\subset \mathbb{R}^{\varrho}$, with the flow set $C_r(\mathbf{Y_c})\subset \mathbb{R}^{\varrho}$, $C_r:C_c\rightrightarrows
\mathbb{R}^{\varrho}$, 
jump set $D_r(\mathbf{Y_c})\subset \mathbb{R}^{\varrho}$, $D_r:C_c\rightrightarrows \mathbb{R}^{\varrho}$, and dynamics
\begin{align}
    \dot{\mathbf{R}} &= f_r(\mathbf{Y_c}, \mathbf{R}), &\mathbf{R}\in C_r(\mathbf{Y_c})\label{eq:flow}, \\
    \mathbf{R}^{+} &= g_r(\mathbf{Y_c}, \mathbf{R}), &\mathbf{R}\in D_r(\mathbf{Y_c}),\label{eq:jump}
\end{align}
with the flow map $f_r:  C_c\times C_r(\mathbf{Y_c})\rightarrow C_r(\mathbf{Y_c})\cup D_r(\mathbf{Y_c})$ and jump map $g_r: C_c\times D_r(\mathbf{Y_c}) \rightarrow C_r(\mathbf{Y_c})\cup D_r(\mathbf{Y_c})$.

\begin{asm}\label{asm:cii_osclb}
The sets $C_r(\mathbf{Y_c})$ and $D_r(\mathbf{Y_c})$ are closed. The function $f_r$ is continuous with respect to $C_c\times C_r(\mathbf{Y_c})$, and $f_r(\mathbf{Y_c},\mathbf{R})$ is jointly convex $\forall$ $\mathbf{Y_c}\in C_c$ and $\mathbf{R}\in C_r(\mathbf{Y_c})$. The function $g_r$ is continuous with respect to $C_c\times D_r(\mathbf{Y_c})$.
\end{asm}

\begin{rem}
As noted in \cite{maass_state_2021},  \eqref{eq:flow}--\eqref{eq:jump} can capture all common interface protocols, for example the zero order hold and first order hold are given by:
\begin{align}
    &f_r^{ZOH}(\mathbf{Y_c},\mathbf{R}) = 0,
    &g_r^{ZOH}(\mathbf{Y_c},\mathbf{R}) = \mathbf{Y_c};
    \label{eq:ciiZOH}\\
    &f_r^{FOH}(\mathbf{Y_c},\mathbf{R}) = \frac{\mathbf{Y_c^k}-\mathbf{Y_c^{k-1}}}{t_k-t_{k-1}},
    &g_r^{FOH}(\mathbf{Y_c},\mathbf{R}) = \mathbf{Y_c},
\end{align}
where $t_{k-1}$ and $t_k$ denote the times of the two most recent jumps.
In \cite{wang_co-design_2017}, the jump map  $\mathbf{Y_c}$ is the desired separation between clusters in two dimensions.

\end{rem}

\subsection{System Controller and Dynamics}

The autonomous system controller admits the reference $\mathbf{R}$ and an environmental input $E_a\in C_a\subset\mathbb{R}^{a}$, and updates its internal state $\mathbf{X}\in C_x\subset \mathbb{R}^{n_p}$ and output $\mathbf{Y_a}\in C_m\subset\mathbb{R}^m$ according to
\begin{align}
    \dot{\mathbf{X}} &= f_a(\mathbf{R},E_a,\mathbf{X}), \label{eq:plantstate}\\
    \mathbf{Y_a} &= h_a(\mathbf{X} ), \label{eq:measuredplantstate}
\end{align}
where the state map $f_a:C_r(\mathbf{Y_c})\cup D_r(\mathbf{Y_c})\times\mathbb{R}^{a}\times C_x\rightarrow C_x$ and the output map $h_a:C_x\rightarrow C_{m}$.

\begin{asm}\label{asm:scd_osclb}
The sets $C_x$ and $C_m$ are closed. 
The function $f_a$ is continuous with respect to $C_r(\mathbf{Y_c})\cup D_r(\mathbf{Y_c})\times C_a\times C_x$, and $f_a(\mathbf{R},E_a,\mathbf{X})$ is jointly convex $\forall$ $\mathbf{R}\in C_r(\mathbf{Y_c})\cup D_r(\mathbf{Y_c})$, $E_a\in C_a$ and $\mathbf{X}\in C_x$. 
The function $h_a$ is continuous and differentiable with respect to $C_x$, and $h_a(\mathbf{X})$ is convex $\forall$ $\mathbf{X}\in C_x$.
\end{asm}

\begin{rem}
The dynamics $f_a$ can incorporate a state feedback controller, for example in \cite{xu_optimo_2015}
\begin{align}
    f_a = \phi(\mathbf{X},E_a)+k_p(\mathbf{X} - \mathbf{R}),
\end{align}
where $\phi:C_x\times C_a\rightarrow C_x$ denotes a boundary-tracking algorithm and $k_p\in\mathbb{R}$ is an error-correcting coefficient.

\end{rem}

\subsection{System Status Interface}
In the proposed framework, the autonomous system output, $\mathbf{Y_a}$, is measured and filtered by the system status interface. The filter output $\mathbf{Y_s}$ is subsequently transmitted to the commander. 
The dynamics of $\mathbf{S}\in C_s(\mathbf{Y_a})\cup D_s(\mathbf{Y_a})$ are given by
\begin{align}
    \dot{\mathbf{S}} &= f_s(\mathbf{Y_a},\mathbf{S}), &\mathbf{S}\in C_s(\mathbf{Y_a})\label{eq:status_flow}, \\
    \mathbf{S}^{+} &= g_s(\mathbf{Y_a}, \mathbf{S}), &\mathbf{S}\in D_s(\mathbf{Y_a}),\label{eq:status_jump}\\
       {\mathbf{Y_s}} &= h_s(\mathbf{S}),\label{eq:status_output}
\end{align}
with the flow map $f_s:  C_{m}\times C_s(\mathbf{Y_a})\rightarrow C_s(\mathbf{Y_a})\cup D_s(\mathbf{Y_a})$, flow set $C_s(\mathbf{Y_a})\subset \mathbb{R}^{s}$ with $C_s:C_m\rightrightarrows\mathbb{R}^s$, jump map $g_s: C_m\times D_s(\mathbf{Y_a}) \rightarrow C_s(\mathbf{Y_a})\cup D_s(\mathbf{Y_a})$, and jump set $D_s(\mathbf{Y_a})\subset \mathbb{R}^{s}$ with $D_s:C_m\rightrightarrows\mathbb{R}^s$. 
The output map $h_s:C_s(\mathbf{Y_a})\cup D_s(\mathbf{Y_a})\rightarrow C_{\sigma}$ yields the interface output $\mathbf{Y_s}\in C_{\sigma}\subset\mathbb{R}^{\sigma}$.

\begin{asm}\label{asm:ssi_osclb}
The sets $C_s(\mathbf{Y_a})$, $D_s(\mathbf{Y_a})$ and $C_\sigma$ are closed.
The function $f_s$ is continuous with respect to $C_m\times C_s(\mathbf{Y_a})$, and $f_s(\mathbf{Y_a},\mathbf{S})$ is jointly convex $\forall$ $\mathbf{Y_a}\in C_m$ and $\mathbf{S}\in C_s(\mathbf{Y_a})$. 
The function $g_s$ is continuous with respect to $C_m\times D_s(\mathbf{Y_a})$. 
The function $h_s$ is continuous and differentiable with respect to $C_s(\mathbf{Y_a})\cup D_s(\mathbf{Y_a})$, and $h_s(\mathbf{S})$ is jointly convex $\forall$ $\mathbf{S}\in C_s(\mathbf{Y_a})\cup D_s(\mathbf{Y_a})$ 
\end{asm}

\begin{rem}
Many existing sampling protocols are described by \eqref{eq:status_flow}--\eqref{eq:status_jump}. For example in \cite{mahani_bayesian_2020} the sampling dynamics are zero order hold measurements described by $f_s = \mathbf{0}$ and $g_s = \mathbf{Y_a}$ where $\mathbf{Y_a}=[v^r,d^r,s^r]_{r\in\{1,N\}}$, $v^r\in[0,1]$ is the normalized robot velocity, $d^r\in[0,1]$ is the ratio of distance traveled to total distance, and $s^r\in[0,1]$ is the robot's detection accuracy.
\eqref{eq:status_output} can convey system states directly to the commander (e.g. $h_s(\mathbf{S}) = \mathbf{S}$ in \cite{liu_clustering_2021}), 
or provide a recommendation to the commander (e.g. in \cite{mcmahon_modeling_2020} $h_s(\mathbf{S}) = \alpha(\mathbf{S}) S_A^- + (1-\alpha(\mathbf{S}) )S_A^+$ where $\alpha(\mathbf{S})\in\{0,1\}$ denotes the presence of a stimulus).
\end{rem}

\subsection{Performance Estimation}
The commander's estimate of the system's performance on a given task $\mathbf{P}\in C_p\subset \mathbb{R}$ is influenced by $\mathbf{Y_s}$ and a trust parameter $\kappa\in C_{k}\subset\mathbb{R}^k$ which captures variations in trust and intervention tendencies between individuals (cf. trust preference polytopes \cite{vella_individual_2022}).
The state update map $f_p:C_{k}\times C_{\sigma}\times C_p\rightarrow C_p$ yields
\begin{align}
    \dot{\mathbf{P}} &= f_p(\kappa, \mathbf{Y_s}, \mathbf{P}).\label{eq:pdot}
\end{align}
\begin{asm}\label{asm:pe_osclb}
    $C_p$ is closed, $f_p$ is continuous with respect to $C_k\times C_\sigma \times C_p$, and $f_p(\kappa, \mathbf{Y_s}, \mathbf{P})$ is jointly convex $\forall$ $\kappa\in C_k$, $\mathbf{Y_s}\in C_\sigma$, $\mathbf{P}\in C_p$.
\end{asm}

\begin{rem}
In \cite{wang_co-design_2017} the authors consider a swarm composed of several clusters of agents; the performance of the $i$th cluster can be implemented as $f_p= (1-k_m)\mathbf{Y_{s,1}} + k_m\mathbf{Y_{s,2}}-\mathbf{P}$, where $\mathbf{Y_{s,1}}\in[0,1]$ represents cluster cohesion, $\mathbf{Y_{s,2}}\in[0,1]$ measures how well the cluster leader maintains inter-cluster distances, and $k_m$ is a tunable parameter.
\eqref{eq:pdot} can be implemented for \cite{nam_predicting_2017} as $f_p=\frac{1}{N}\mathbf{Y_s}$ with $\mathbf{Y_s}$ denoting the instantaneous rate of finding $N$ goals, and for \cite{lee_trust_1992} as $f_p = \frac{\mathbf{Y_s}}{r_i}-\mathbf{P}$ where $r_i$ is the input flow rate and $\mathbf{Y_s}$ is the output flow rate.
\end{rem}

\subsection{Commander Trust and Intervention Dynamics}

The trust dynamics admit as inputs $\mathbf{Y_s}$, $\mathbf{P}$, and an environmental input $E_c\in C_b\subset \mathbb{R}^b$ (which is not necessarily identical to $E_a$).
Trust $T\in C_t\subset \mathbb{R}$ is given by
\begin{align}
    \dot{\mathbf{T}} = f_c(\kappa, \mathbf{P}, E_c, \mathbf{T}).\label{eq:trustdyns_ol}
\end{align}
with the individual nature of the trust state update map $f_c:C_k \times C_p  \times C_b \times C_t \rightarrow C_t$ again captured through the inclusion of the parameter $\kappa$.

\begin{asm}\label{asm:ctid_osclb}
    $C_t$ is closed, $f_c$ is continuous with respect to $C_k\times C_p\times C_b\times C_t$, and $f_c(\kappa, \mathbf{P}, E_c, \mathbf{T})$ is jointly convex $\forall$ $\kappa\in C_k$, $\mathbf{P}\in C_p$, $E_c\in C_b$ and $\mathbf{T}\in C_t$.
\end{asm}

\begin{rem}
Linear system models have been proposed for $f_c$ in \eqref{eq:trustdyns_ol} (e.g. state space models of the form
$f_c = A \mathbf{T} + B\mathbf{P}$ given in \cite{lee_trust_1992,akash_dynamic_2017}), however linear dynamics cannot describe situations where trust responds asymmetrically to changes in performance.
The \textit{quick to lose, slow to gain} nature of trust has been observed in 
\cite{lee_trust_1992,akash_dynamic_2017,liu_clustering_2021}, with faults triggering a sudden decrease and slow recovery in trust. 
Nonlinear dynamics that model this phenomenon are compatible with \eqref{eq:trustdyns_ol}.
\end{rem}

Driven by $\mathbf{T}$, the commander issues an intervention $\mathbf{Y_c}\in C_c \subset \mathbb{R}^c$ to improve system performance.
The output map $h_c:C_k\times C_{\sigma} \times C_t \rightarrow C_c$ yields
\begin{align}
    \mathbf{Y_c} = h_c(\kappa,\mathbf{Y_s},\mathbf{T}).\label{eq:CTID_output}
\end{align}

\begin{asm}\label{asm:ctid_osclb2}
    $C_c$ is closed, $h_c$ is continuously differentiable with respect to $C_k\times C_\sigma\times C_t$, and
    $h_c(\kappa,\mathbf{Y_s},\mathbf{T})$ is jointly convex $\forall$ $\kappa\in C_k$, $\mathbf{Y_s}\in C_\sigma$, $\mathbf{T}\in C_t$.
\end{asm}
\begin{rem}
    In \cite{mcmahon_modeling_2020} \eqref{eq:CTID_output} can be implemented as $h_c = \mathcal{E}(\mathbf{T})$, where $\mathcal{E}:C_t\rightarrow C_c$ maps trust to the commander's decision to accept the system's advice.
    In \cite{mahani_bayesian_2020} trust influences when the commander switches to a manual detection mode and can be implemented as $h_c = \psi(\mathbf{Y_s},\mathbf{T})$, where $\psi:C_\sigma\times C_t\rightarrow C_c$ captures (7) in \cite{mahani_bayesian_2020}. 
\end{rem}

\section{State Convergence under Event-Triggered Sampling}\label{sec:convAETS}

We now analyze the network of subsystems in Figure \ref{fig:framework} using an approach motivated by \cite{williams_asynchronous_2024}, in which the system status interface acts as a sampler for the autonomous system and the commander intervention interface is a sampler for the commander-related subsystems.

\subsection{Plant Dynamics under ETS}

For the autonomous system and the system status interface, we define the plant offset $x_p := \mathbf{X}-\mathbf{X}^{*} \in\mathbb{R}^{n_p}$ where $\mathbf{X}^{*}\in C_x$ is a state manifold for the autonomous system, the plant output $x_m := h_a(x_p)\in\mathbb{R}^{n_m}$, the plant output sampler offset $\hat{x}_m := \mathbf{S}-\mathbf{S}^*\in \mathbb{R}^{n_m}$ where $\mathbf{S}^*\in \mathbb{R}^{n_m}$ is a state manifold for the system status interface, and the plant output sampling error $e_m := \hat{x}_m-x_m\in \mathbb{R}^{n_m}$.
An auxiliary variable $\eta_p\in\mathbb{R}_{\geq0}$ records time elapsed since the last plant sample with
$\dot{\eta}_p = 1$ and $\eta_p^+ = 0$.
We use $\eta_p$ in the sampler's trigger condition to impose a minimum sampling interval $\tau_p>0$ (preventing Zeno behavior when triggering new samples \cite{abdelrahim_robust_2017}).
Combining the plant-related variables as $q_p = [x_p^T, e_m^T, \eta_p^T]^T\in\mathbb{R}^{n_{q,p}}$, we then select a locally Lipschitz function $V_p:\mathbb{R}^{n_{q,p}}\rightarrow\mathbb{R}_{\geq0}$ and a continuously differentiable function $W_p:\mathbb{R}^{n_m}\rightarrow\mathbb{R}_{\geq0}$ so that sampling is triggered when both $V_p(q_p)\leq W_p(e_m)$ and $\eta_p=\tau_p$.
To this end we define the flow set
 \begin{align}
C_s := \{q_p: V_p(q_p)>W_p(e_m)\,\&\,\eta_p<\tau_{p}\}\label{eq:plocalflowset}
\end{align}
the jump set
\begin{align}
D_s := \{q_p: V_p(q_p)\leq W_p(e_m)\,\&\,\eta_p=\tau_p\}\label{eq:plocaljumpset}
\end{align}
and the set $\mathcal{A}_p=\{q_p:x_p = 0, e_p=0, \eta_p\in\mathbb{R}\}$.

\subsection{Controller Dynamics under ETS}
Similarly for the commander susbsystems and the commander intervention interface, we define the controller offsets $x_c = [x_{c,1}^T, x_{c,2}^T]^T \in C_p\times C_t \subseteq \mathbb{R}^{n_c}$ such that $x_{c,1} = (\mathbf{P}-\mathbf{P}^{*})\in C_p$ and $x_{c,2} = (\mathbf{T}-\mathbf{T}^{*})\in C_t$,
the controller output $u = \mathbf{Y_c} - \mathbf{Y_c^*} \in C_c\subseteq \mathbb{R}^{n_u}$ and an arbitrary intervention output equilibrium $\mathbf{Y_c^*}\in C_c$, the controller sampler offset $\hat{u} = \mathbf{R}-\mathbf{R}^*\in\mathbb{R}^{n_u}$ and an arbitrary intervention state manifold $\mathbf{R}^*\in\mathbb{R}^{n_u}$, and the controller sampling error $e_u:=\hat{u} - u\in\mathbb{R}^{n_u}$. 
An auxiliary variable $\eta_c\in\mathbb{R}_{\geq0}$ records time elapsed since the last controller sample with
$\dot{\eta}_c =1$ and $\eta_c^+ = 0$.
Similar to $\eta_p$, we use $\eta_c$ to impose a minimum sampling interval $\tau_c > 0$ (preventing Zeno behavior).
Defining the aggregate variable $q_c = [x_c^T, e_u^T, \eta_c^T]^T\in\mathbb{R}^{n_{q,c}}$, we choose a locally Lipschitz function $V_c:\mathbb{R}^{n_{q,c}}\times\mathbb{R}_{\geq0}\rightarrow\mathbb{R}_{\geq0}$ and a continuously differentiable function $W_u:\mathbb{R}^{n_u}\rightarrow\mathbb{R}_{\geq0}$ so that sampling is triggered when both $V_c(q_c)\leq W_u(e_u)$ and $\eta_c=\tau_c$.
We define the local flow set
\begin{align}
    C_r &:= \{q_c: V_c(q_c)> W_u(e_u)\,\&\,\eta_c<\tau_{c}\},\label{eq:clocalflowset}\end{align}
jump set \begin{align}
    D_r &:= \{q_c: V_c(q_c)\leq W_u(e_u)\,\&\,\eta_c=\tau_c\},\label{eq:clocaljumpset}
\end{align}
and the set $\mathcal{A}_c=\{q_c:x_c = 0, e_u=0, \eta_c\in\mathbb{R}\}$.

\subsection{Closed-Loop Dynamics under ETS}
For the closed-loop system we consider the flow set 
\begin{align} 
C&=\{(q_p,q_c)\in\mathbb{R}^{n_p+n_c}:q_p\in C_s\;\&\;q_c\in C_r\}\end{align}
and the jump set 
\begin{align} 
D&=\{(q_p,q_c)\in\mathbb{R}^{n_p+n_c}:q_p\in D_s\,\parallel\,q_c\in D_r\}.\end{align}
Let $\hat{x}_p := x_p + e_p$ and $\hat{u} := u + e_u$.
Substituting $\mathbf{X}=x_p+\mathbf{X^*}$, $\mathbf{R}=\hat{u}+\mathbf{R^*}$, $\mathbf{Y_a} = h_a(\mathbf{X})$, $\mathbf{Y_s} = h_s(\hat{x}_m+\mathbf{S^*})$, $\mathbf{P}=x_{c,1}+\mathbf{P^*}$, $\mathbf{T} = x_{c,2}+\mathbf{T^*}$, and $\mathbf{Y_c} = h_c(\kappa, \mathbf{Y_s},\mathbf{T})$, the flow maps for $q_p$ and $q_c$ can be rewritten as
\begin{align}
F_1(q_p,q_c)&=\begin{bmatrix}
f_a(\mathbf{R}, E_a,\mathbf{X})\\ 
 f_s(\mathbf{Y_a}, \mathbf{S}) - \frac{dh_a(x_p)}{dx_p}f_a(\mathbf{R},E_a,\mathbf{X})\\ 
1\end{bmatrix},\label{eq:F1}\\
F_2(q_p,q_c)&=\begin{bmatrix}
    \phi_1(q_p,q_c)\\
    \phi_2(q_p,q_c)\\
    1
\end{bmatrix},\label{eq:F2}
\end{align}
where
    $\phi_1(q_p,q_c) = \begin{bmatrix}
        f_p(\kappa, \mathbf{Y_s},\mathbf{P})\\
        f_c(\kappa, \mathbf{P}, E_c, \mathbf{T})
    \end{bmatrix}$
and
$
    \phi_2(q_p,q_c) = \dot{\hat{u}}-\dot{u} = \mathbf{\dot{R}} -\mathbf{\dot{Y}_c}= f_r(\mathbf{Y_c},\mathbf{R})
 -\frac{\partial\mathbf{Y_c}}{\partial\mathbf{Y_s}} \frac{d\mathbf{Y_s}}{d\mathbf{S}} \mathbf{\dot{S}}
-\frac{\partial\mathbf{Y_c}}{\partial\mathbf{T}}\mathbf{\dot{T}}.$
The respective jump maps for $q_p$ and $q_c$ can be rewritten as
\begin{align}
    G_1(q_p,q_c) &= \begin{cases}
    \begin{bmatrix}
        x_p^T,0^T,0^T
    \end{bmatrix}^T, &(q_p,q_c)\in D_s\times C_r,\\
    \begin{bmatrix}
        x_p^T,e_m^T, \eta_p^T
    \end{bmatrix}^T, &(q_p,q_c)\in C_s\times D_r,\\
    \left\{\begin{bmatrix}
        x_p\\0\\0
    \end{bmatrix},\begin{bmatrix}
        x_p\\e_m\\ \eta_p
    \end{bmatrix}\right\}, &(q_p,q_c)\in D_s\times D_r, \label{eq:pdoublejump}
    \end{cases}
\end{align}
\begin{align}
    G_2(q_p,q_c) &= \begin{cases}
    \begin{bmatrix}
        x_c^T, e_u^T, \eta_c^T
    \end{bmatrix}^T, &(q_p,q_c)\in D_s\times C_r,\\
    \begin{bmatrix}
        x_c^T,0^T,0^T
    \end{bmatrix}^T, &(q_p,q_c)\in C_s\times D_r,\\
    \left\{\begin{bmatrix}
        x_c\\e_u\\\eta_c
    \end{bmatrix},\begin{bmatrix}
        x_c\\0\\0
    \end{bmatrix}
    \right\}, &(q_p,q_c)\in D_s\times D_r. \label{eq:cdoublejump}
    \end{cases}
\end{align}
Note that \eqref{eq:pdoublejump} and \eqref{eq:cdoublejump} denote successive jumps when both samplers trigger simultaneously; this ensures that the jump maps remain outer-semi-continuous \cite{abdelrahim_robust_2017}.
The hybrid dynamics of the closed-loop system state $q = [q_p^T, q_c^T]^T$ can thus be represented with\begin{align}
    \dot{q} &\in F(q), &q\in C,\label{eq:sysflows} \\
    q^+ &\in G(q), &q\in D, \label{eq:sysjumps}
\end{align}
where $F(q) = [F_1(q_p,q_c)^T, F_2(q_p,q_c)^T]^T$ and $G(q) = [G_1(q_p,q_c)^T, G_2(q_p,q_c)^T]^T$.

We next adapt \cite[Assumption III.2]{liberzon_lyapunov-based_2014} in order to later use the hybrid small-gain theorem in \cite{liberzon_lyapunov-based_2014}.
To do so we first introduce two definitions.

\begin{defn}
    The tangent cone $T_S(x)$ to a set $S\subset\mathbb{R}^n$ at a point $x\in\mathbb{R}^n$ is the set $\{w\in\mathbb{R}^n: w = \lim_{i\rightarrow\infty}\frac{x_i-x}{\tau_i}\}$ where $x_i\in S$ is such that $\lim_{i\rightarrow\infty}x_i = x$ and $\tau_i>0$ is such that $\lim_{i\rightarrow\infty}\tau_i = 0$.
\end{defn}
\begin{defn}
The generalized Clarke directional derivative of a locally Lipschitz function $U:\mathbb{R}^n\rightarrow\mathbb{R}$ at $x\in\mathbb{R}^n$ in the direction of $v\in\mathbb{R}^n$ is given by  $U^{\circ}(x;v) = \limsup_{h\rightarrow 0^+,y\rightarrow x}(U(y+hv)-U(y))/h$.
\end{defn}

\begin{asm}\label{asm:smallgain}
The functions $V_p$ and $V_c$ used in \eqref{eq:plocalflowset}--\eqref{eq:plocaljumpset} and \eqref{eq:clocalflowset}--\eqref{eq:clocaljumpset} satisfy the following:
\begin{enumerate}
    \item \label{asm:smallgain1} there exist functions $\underline{\alpha}_p,\bar{\alpha}_p,\underline{\alpha}_c,\bar{\alpha}_c\in\mathcal{K}_{\infty}$ such that $\forall$ $q_p\in\mathbb{R}^{n_{q,p}}$ and $q_c\in\mathbb{R}^{n_{q,c}}$,
    \begin{align}
        &\underline{\alpha}_p(|q_p|_{\mathcal{A}_p}) \leq V_p(q_p) \leq \bar{\alpha}_p(|q_p|_{\mathcal{A}_p}),\\
        &\underline{\alpha}_c(|q_c|_{\mathcal{A}_c}) \leq V_c(q_c) \leq \bar{\alpha}_c(|q_c|_{\mathcal{A}_c});
    \end{align} 
    \item \label{asm:smallgain2} there exist functions $\chi_p,\chi_c\in\mathcal{K}_{\infty}\cup\{0\}$, $\alpha_p,\alpha_c:\mathbb{R}_{\geq 0}\rightarrow \mathbb{R}_{\geq 0}$ such that $\forall$ $(q_p,q_c)\in C$,
    \begin{align}
        V_p(q_p)\geq \chi_p(V_c(q_c))\implies& \nonumber\\
        V_p^{\circ}(q_p,F_1(q_p,q_c))\leq -\alpha_p(|q_p|_{\mathcal{A}_p}),&\\
        V_c(q_c)\geq \chi_c(V_p(q_p))\implies& \nonumber\\
        V_c^{\circ}(q_c,F_2(q_p,q_c))\leq -\alpha_c(|q_c|_{\mathcal{A}_c});&
    \end{align}
    \item \label{asm:smallgain3} $\forall$ $(q_p,q_c)\in D$, 
    \begin{align}
        V_p(G_1(q_p,q_c))\leq V_p(q_p),\\
        V_c(G_2(q_p,q_c))\leq V_c(q_c);
    \end{align}
    \item \label{asm:smallgain4} $\chi_p \circ \chi_c (s) < s $ $\forall$ $s>0$;
    \item \label{asm:smallgain5} there exists a function $\rho\in\mathcal{K}_{\infty}$ that is continuously differentiable on $(0,\infty)$ satisfying both $\chi_p(r)<\rho(r)<\chi_c^{-1}(r)$ and $\rho'(r)>0$ $\forall$ $r>0$.
\end{enumerate}
\end{asm}

\begin{rem}\label{rem:hard}
In the context of \eqref{eq:sysflows}--\eqref{eq:sysjumps}, the conditions of Assumption \ref{asm:smallgain} are conservative and imply that the growth in the error variables $e_p$ and $e_u$ depends on the choice of controller and plant gains, and the minimum sampling periods $\tau_p$ and $\tau_c$.
For all but the most trivial systems, it is challenging to analytically determine the functions necessary to satisfy Assumption \ref{asm:smallgain}, however we may use the intuition underlying these conditions to choose parameter for more complex systems.
\end{rem}

The following adapted from \cite[Theorem 4.1]{williams_asynchronous_2024} concerns the stability of $\mathcal{A}_p\times\mathcal{A}_c$ 
for solutions of \eqref{eq:sysflows}--\eqref{eq:sysjumps}. 

\begin{thm}\label{thm:ugasA}
Given the system \eqref{eq:sysflows}--\eqref{eq:sysjumps} under Assumptions \ref{asm:cii_osclb}--\ref{asm:smallgain}, if there exists a tuple of positive constants $(\tau^*_p,\tau^*_c)$ such that for all $\tau_p\in(0,\tau^*_p)$ and $\tau_c\in(0,\tau^*_c),$ $G(D)\subset C\cup D$ and $F(q)\in T_C(q)$ for any $q\in C\backslash D$, then $\mathcal{A}_p\times\mathcal{A}_c$ is uniformly globally asymptotically stable.
\end{thm}

\begin{IEEEproof}
Let $\phi$ denote a solution of the system \eqref{eq:sysflows}--\eqref{eq:sysjumps} under Assumption \ref{asm:smallgain}. 
Define a storage function $U(q) = \max\left(V_p(q_p),\rho(V_c(q_c))\right)$. 
By \cite[Thm. III.3]{liberzon_lyapunov-based_2014}, 
\begin{enumerate}
    \item there exist functions $\underline{\alpha}_U, \bar{\alpha}_U\in \mathcal{K}_{\infty}$ such that $\forall$ $q\in\mathbb{R}_{n_q}$, $\underline{\alpha}_U(|q|_{\mathcal{A}})\leq U(q)\leq \bar{\alpha}_U(|q|_{\mathcal{A}})$; \label{eq:libthm3.1}
    \item  there exists a positive definite function $\alpha_U:\mathbb{R}_{\geq 0}\rightarrow \mathbb{R}_{\geq 0}$ such that $\forall$ $q\in C \backslash \mathcal{A}$,
        $U^{\circ}(q,F(q))\leq -\alpha_U(|q|_{\mathcal{A}})$;%
        \label{eq:libthm3.2}
    \item $\forall$ $q\in D$, 
        $U(G(q))\leq U(q)$. %
        \label{eq:libthm3.3}
\end{enumerate}

Due to the structure of the jump sets \eqref{eq:plocaljumpset} and \eqref{eq:clocaljumpset}, a continuous time interval of at least $\tau_p$ exists between any pair of jumps in $q_p$; these jumps are caused either by the system status interface alone (denoted by `$p$') or by both interfaces (`$b$'). 
Similarly a continuous time interval of at least $\tau_c$ elapses between any pair of jumps in $q_c$ caused by both samplers (`$b$') or the commander intervention interface alone (`$c$'). 
If any sequence of at least three jumps occurs, then at least two of these jumps involve the same state (e.g. the sequence `$cpc$' involves two jumps in $q_c$, while the sequence `$bc$' involves two jumps in $q_p$. 
We can conclude that an interval of at least $\underline{\tau}:=\min(\tau_p,\tau_c)$ must elapse between the first and the third jump.
Define the quantity $\tau_a = \frac{\underline{\tau}}{2}$.
We recall from \cite{goebel_hybrid_2012} that the solution's hybrid time domain $\mathrm{dom}\phi$ satisfies $\mathrm{dom}\phi \cap ([0,t]\times\{0,...,j\}) = \bigcup_{i\in\{0,...,j\}}[t_i,t_{i+1}]\times\{i\}$
for any point $(t,j)\in \mathrm{dom}\phi$ and sequence of times $0=t_0\leq t_1\leq...\leq t_{j+1}=t$.
Given $(s,i),(t,j)\in\mathrm{dom}\phi$, we wish to show that if $(s+i)\leq(t+j)$, then
\begin{align}
    j-i\leq\frac{t-s}{\tau_a}+1 \label{eq:quasidwelltime}
\end{align}
holds $\forall$ $(j-i)\in\mathbb{N}\cup\{0\}$.
If $j-i=0$, $(s+i)\leq(t+j)$ reduces to $s\leq t$; since $\tau_a>0$, \eqref{eq:quasidwelltime} holds.
If $j-i = 1$,  $(s+i)\leq(t+j)$ reduces to $s\leq t+1$, however by the definition of $\mathrm{dom}\phi$ necessarily $s\leq t$ for $i<j$, so \eqref{eq:quasidwelltime} holds.
If $j-i = 2n$, $n\in\mathbb{N}$, then we apply the result that $t-s \geq n\underline{\tau}=2n\tau_a$ to show that \eqref{eq:quasidwelltime} holds.
If $j-i = 2n+1$, $n\in\mathbb{N}$, we again apply $t-s \geq 2n\tau_a$ to show that \eqref{eq:quasidwelltime} holds $\forall$ $(j-i)\in\mathbb{N}\cup\{0\}$.
Given \eqref{eq:quasidwelltime} and taking $(s+i)=0$, $(t+j)\geq 0$ implies 
\begin{align}
    j\leq \frac{t}{\tau_a}+1. \label{eq:jumptimebound}
\end{align}
Applying \eqref{eq:jumptimebound} to $(t+j)\geq T$ for some $T>0$ yields
\begin{align}
    t\geq \gamma_r(T) - N_r, \label{eq:timebound}
\end{align}
where $\gamma_r(s) = (1+\tau_a^{-1})^{-1}s$ for $s>0$, and $N_r=(1+\tau_a^{-1})^{-1}$.
Given \eqref{eq:libthm3.1}--\eqref{eq:libthm3.3} and \eqref{eq:timebound}, we now apply \cite[Prop. 3.27]{goebel_hybrid_2012} to demonstrate that the set $\mathcal{A}_p\times\mathcal{A}_c$ is uniformly globally pre-asymptotically stable.
Since $\mathcal{A}_p\times\mathcal{A}_c$ is compact, following \cite{postoyan_framework_2015} we can use \cite[Prop. 6.10]{goebel_hybrid_2012} to show that maximal solutions of \eqref{eq:sysflows}--\eqref{eq:sysjumps} are complete.
Hence by \cite[Def. 4]{postoyan_framework_2015}, $\mathcal{A}_p\times\mathcal{A}_c$ is uniformly globally asymptotically stable. 
\end{IEEEproof}

\begin{rem}
    Because the sampling scheme is error-driven, intuitively one would expect that systems with faster flow dynamics $(\dot{x}_p,\dot{x}_c)$ require more frequent triggering for the same trigger policy given by \eqref{eq:plocaljumpset}, \eqref{eq:clocaljumpset}. 
\end{rem}

\section{Numerical Example}\label{sec:numel}

\begin{figure}[t]
    \centering
\includegraphics[width=0.75\columnwidth]{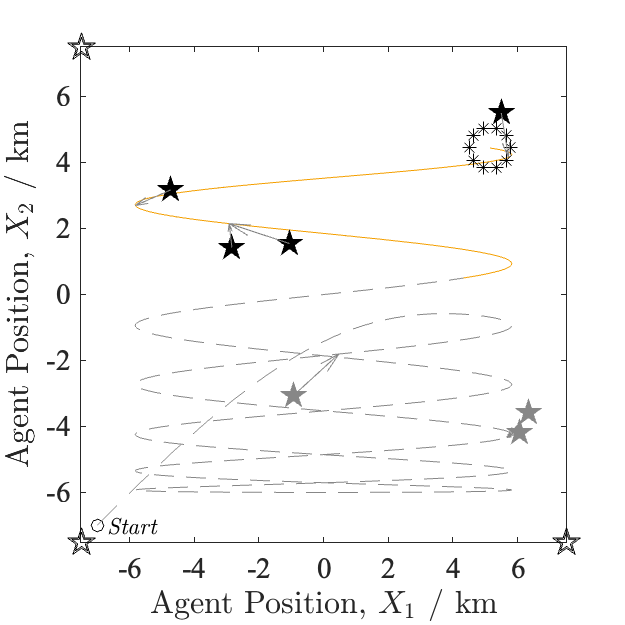}
  \label{fig:m1_it}
  \caption{Swarm centroid trajectory from $t = 0.0$ s to $t=40.0$ s in Mission $M^C_1$. \textit{Legend:} \ding{72} survivor detected from $t=32.5$ s to $t=40.0$ s, \textcolor{grey}{\ding{72}} survivor detected before $32.5$ s, \ding{73} undetected survivor, \ding{213} swarm centroid position when the adjoined survivor was detected, $\circ$ swarm centroid initial position, $-\,-$ trajectory before $t=32.5$ s, 
  \textcolor{gold}{---} trajectory from $t=32.5$ s to $t=40.0$ s, \ding{83} agent positions at $t=40.0$ s.}
  \label{fig:cohesiontrust} 
\end{figure} 

We present an overview of a search and rescue task conducted using the ETS framework described in $\S$\ref{sec:convAETS}.
A human commander supervises an autonomous robotic swarm system consisting of $n_a=10$ agents collectively searching for $n_s=10$ survivors, whose positions $\mathbf{X}_s$ are unknown \textit{a priori} to the commander.
The agents maintain a circular formation defined solely by a commander-determined radius, through the region depicted in Figure \ref{fig:cohesiontrust}.
When agents are sufficiently close to a survivor's position, the system's confidence in having detected that survivor increases, i.e. a smaller formation radius promotes better survivor detection.
This invokes a trade-off, as the formation can cover more search area using a larger radius.
To capture this trade-off, the task performance should reward both survivor detection and maintaining inter-agent proximity.
Commander trust evolves dynamically in response to task performance, which in turn guides the commander's change to formation radius.

\subsection{Implementation}\label{sec:climp}
We now define each of the subsystems for the task;
for clarity we omit unnecessary time indices.

\paragraph{Commander Intervention Interface}
This subsystem generates a reference signal for the formation radius using \eqref{eq:ciiZOH} by sampling the commander's desired agent separation $\mathbf{Y_c}\in[0,1.5]$. This satisfies Assumption \ref{asm:cii_osclb}.
To establish the trigger policy, we define the flow set $C_r(\mathbf{Y_c}) = \{\mathbf{Y_c}: V_c(q_c)> W_u(e_u)\parallel V_c(q_c)\leq W_u(e_u)\;\&\;\eta_c<\tau_{c}\}$,
the jump set $D_r(\mathbf{Y_c}) = \{\mathbf{Y_c}: V_c(q_c)\leq W_u(e_u)\;\&\;\eta_c=\tau_c\}$, 
the state Lyapunov function $V_c(q_c)=(\mathbf{Y_c}-1.41)^2+0.01e_u^2$, 
and the error function $W_u(e_u) = 10.01e_u^2$. 
This combination satisfies Assumption \ref{asm:smallgain}.

\paragraph{System Controller and Dynamics}
The agents' motion is modeled as a single-integrator system 
with
\begin{align}
    \mathbf{\dot{X}}^i =& 4(\mathbf{X}_{\mathbf{ref}}^i-\mathbf{X}^i),\label{eq:X_impl}\\
    \mathbf{X}_{\mathbf{ref}}^i =& 
    \begin{bmatrix}
        6\sin (\frac{t}{2\pi})+\mathbf{R}\cos(\theta_i)\\
        6\sin(\frac{0.1t}{2\pi}) +\mathbf{R}\sin(\theta_i)
    \end{bmatrix}\label{eq:Xr_impl}
\end{align}
where $\theta_i = \frac{2\pi (i-1)}{n_a}$, $i\in\{1,...,n_a\}$ is the bearing of the $i$th agent from the formation centroid.

The system output is a tuple containing a smooth measure of each survivor's proximity to the agents, and swarm cohesion measured by the mean proportion of agents within 1 km of each agent, i.e.
\begin{align}
    \mathbf{Y_a} &= \begin{bmatrix}
    [1-\tanh(\sum_{i=1}^{n_a}\sigma(||\mathbf{X}_s^j-\mathbf{X}^i||_2))]_{j=1}^{n_s}\\
    \frac{\sum_{j=1}^{n_a} (-1+\sum_{i=1}^{n_a}\sigma(||\mathbf{X}^j-\mathbf{X}^i||_2))}{n_a(n_a-1)}
    \end{bmatrix},\label{eq:Ya_impl}
\end{align}
where $\sigma(x)=0.5(1-\tanh(3(x-1)))$ is a smooth indicator.
Assumption \ref{asm:scd_osclb} is satisfied by \eqref{eq:X_impl}--\eqref{eq:Ya_impl}.

\paragraph{System Status Interface}

Let $\mathbf{S_1}(t)\in[0,1]^{n_s}$ represent the likelihood of not finding each survivor before time $t$, such that $\mathbf{S_1}(0)=\mathbf{1}_{n_s}$.
$\mathbf{Y_a}$ is sampled with
\begin{align}
    \mathbf{\dot{S}}(t) &= \mathbf{0},&\mathbf{Y_a}\in C_s(\mathbf{Y_a}),\label{eq:Sdot_impl}\\
    \mathbf{S^+}(t) &= \begin{bmatrix}
        \mathbf{S_1}(t) &0\\
        0 &1
    \end{bmatrix} \mathbf{Y_a}&\mathbf{Y_a}\in D_s(\mathbf{Y_a}).\label{eq:Sj_impl}
\end{align}

For the trigger policy we define the flow set $C_s(\mathbf{Y_a}) = \{\mathbf{Y_a}: V_p(q_p)> W_p(e_m)\parallel V_p(q_p)\leq W_p(e_m)\;\&\;\eta_p<\tau_{p}\}$,
the jump set $D_s(\mathbf{Y_a}) = \{\mathbf{Y_a}: V_p(q_p)\leq W(e_m)\;\&\;\eta_p= \tau_p\}$,
the state Lyapunov function $V_p(q_p)= \sum_{i=1}^{n_s}((\mathbf{Y}_{\mathbf{a},i}^1)^2+0.01\sum_{i=1}^{n_s} (e_{m,i})^2$, 
and the error function $W_p(e_m) = 10.01\sum_{i=1}^{n_s} (e_{m,i})^2$.
This combination satisfies Assumption \ref{asm:smallgain}.

The sampler's output relays the proportion of found survivors and swarm cohesion, i.e. $\mathbf{Y_s} = \begin{bmatrix}\frac{-1}{n_s}\sum_{i=1}^{n_s} \mathbf{S}_{1,i} &\mathbf{S_2}\end{bmatrix}^T$.
This together with \eqref{eq:Sdot_impl}--\eqref{eq:Sj_impl} satisfies Assumption \ref{asm:ssi_osclb}.

\paragraph{Performance Estimation}

Inspired by \cite{wang_co-design_2017} and \cite{nam_predicting_2017}, the performance dynamics are driven by a weighted average of the sampler outputs such that
\begin{align}
    \dot{\mathbf{P}} &= \begin{bmatrix}
        0.9 &0.1
    \end{bmatrix}\mathbf{Y_{s}} -\mathbf{P},
\end{align}
which satisfies Assumption \ref{asm:pe_osclb}.

\paragraph{Commander Trust and Intervention Dynamics}

We propose the commander trust dynamic model 
\begin{align}
    \dot{\mathbf{T}} = 0.5 (\mathbf{P}-\mathbf{T}),
\end{align}
The formation radius reference is given by $\mathbf{Y_c}=1.5\mathbf{T}$, thus  Assumptions \ref{asm:ctid_osclb} and \ref{asm:ctid_osclb2} are satisfied.

\begin{table} [b]
\centering
\caption{Simulation Initial Conditions \label{tab:ICs}}
\begin{tabular}{|c|c|c|c|c|}
\hline
\textbf{Mission} & $\mathbf{P_0}$ & $\mathbf{T_0}$ &  $\mathbf{X_0}$ \\ \hline
$M^A_\tau$                & 0.01        & 0.01        &  $3\times \mathbf{1}_{10\times2}$         \\ \hline
$M^B_\tau$                & 0.50         & 0.50         &  $0\times \mathbf{1}_{10\times2}$           \\ \hline
$M^C_\tau$                & 0.99        & 0.99        &  $-7\times \mathbf{1}_{10\times2}$         \\ \hline
\end{tabular}
\end{table}

\begin{figure}[t]
    \centering
    \includegraphics[width=\columnwidth]{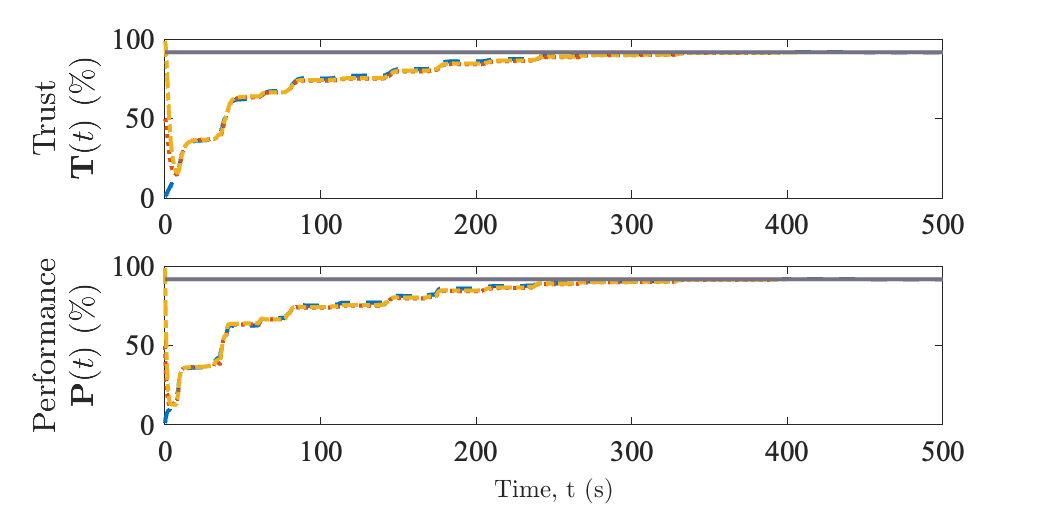}
  \caption{State trajectories in Missions $M^A_{1}$ (blue dashed), $M^B_{1}$ (yellow dash-dotted),  $M^C_{1}$ (red dotted) and $P^*$ or $T^*$  (black solid).}
  \label{fig:CLmissions_1} 
\end{figure} 

\begin{figure}[t]
    \centering
    \includegraphics[width=\columnwidth]{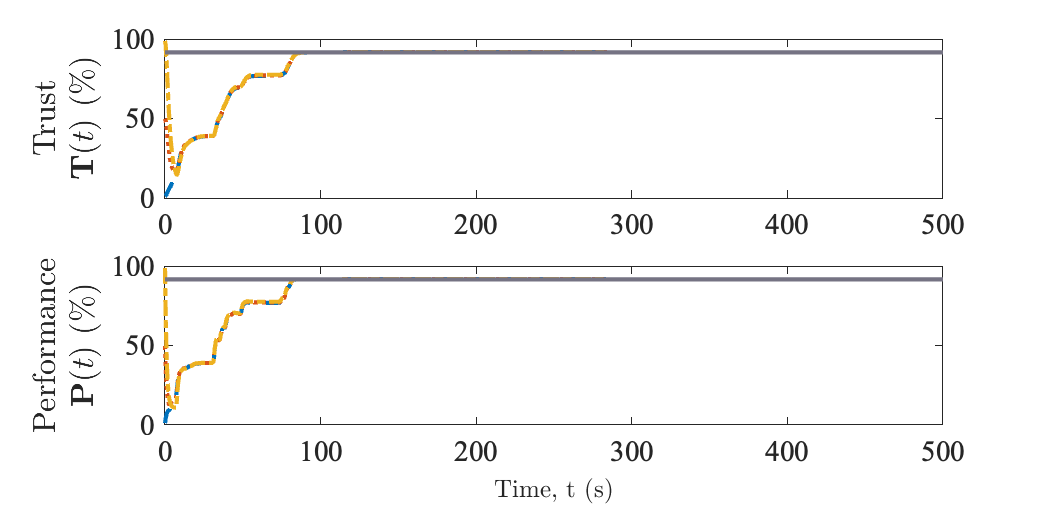}
  \caption{State trajectories in Missions $M^A_{0.1}$
  (blue dashed), $M^B_{0.1}$ (yellow dash-dotted), $M^C_{0.1}$ (red dotted) and $P^*$ or $T^*$  (black solid).}
  \label{fig:CLmissions_0pt1} 
\end{figure} 

\begin{figure}[t]
    \centering
    \includegraphics[width=\columnwidth]{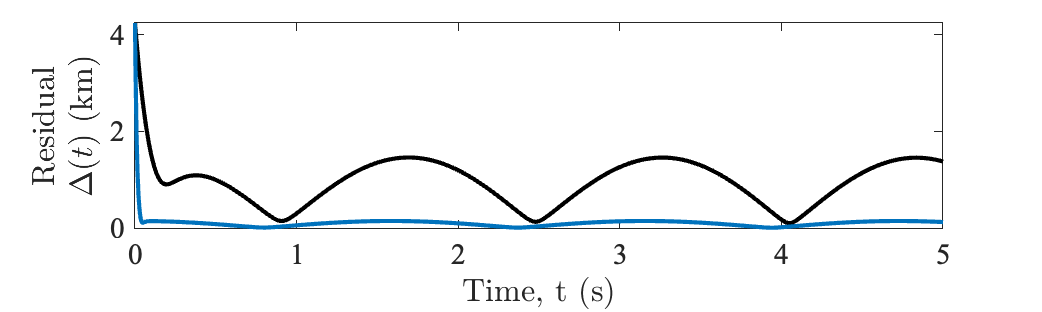}
    \caption{Plant sampler state residual during first 10 seconds of $M_1^A$ for $k_p =4$ (black) and $k_p=40$ (blue).}
  \label{fig:CLM1residuals} 
\end{figure} 

\begin{figure}[t]
    \centering
    \includegraphics[width=\columnwidth]{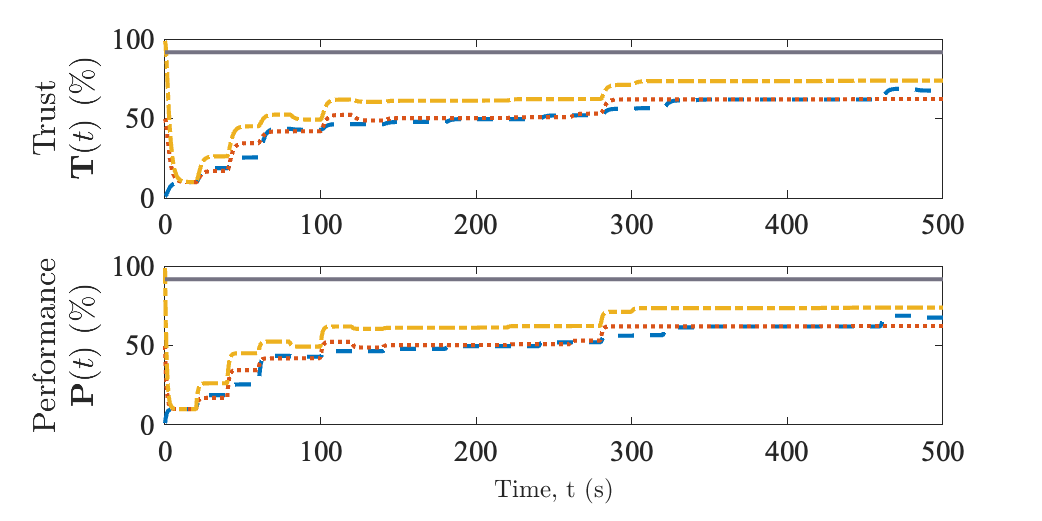}
  \caption{State trajectories in Missions $M^A_{20}$
  (blue dashed), $M^B_{20}$ (yellow dash-dotted), $M^C_{20}$ (red dotted) and $P^*$ or $T^*$  (black solid).}
  \label{fig:CLmissions_20} 
\end{figure}

\vspace{-3mm}
\subsection{Results}

We present results from three simulated search and rescue missions $\{M^A_\tau,M^B_\tau,M^C_\tau\}$ with $\tau=\tau_p=\tau_c$ and the initial conditions included in Table \ref{tab:ICs}.
As illustrated in Figure \ref{fig:cohesiontrust} for $M^B_1$, the agents traverse the search region and report the locations of detected survivors to the commander.
The commander is modeled such that an increase in system performance causes an increase in trust, which in turn leads to an increased formation radius (conversely a decrease in performance leads to decreases in trust and the formation radius).

We first observe how the asymptotic behavior of the framework's subsystems is influenced by the choice of the parameters.
Recall that $q_c=[x_c,e_u,\eta_c]^T$, where $x_c=[\mathbf{P}-\mathbf{P^*},\mathbf{T}-\mathbf{T^*}]^T$.
Figures \ref{fig:CLmissions_1} and \ref{fig:CLmissions_0pt1} illustrate how $q_c$ converges asymptotically to the same (global) equilibrium $\mathcal{A}_c$, with the rate of convergence dependent on the magnitudes of $\tau_p$ and $\tau_c$. 

Next to facilitate an investigation of the convergence towards $\mathcal{A}_p$, we introduce the residual   $\mathbf{\Delta}(t) := \left\|\frac{\sum_{i=1}^{n_a} \mathbf{X}^i(t)-\mathbf{X}_{\mathbf{ref}}^i(t)}{n_a}\right\|_2$.
Figure \ref{fig:CLM1residuals} depicts the residual for the first ten seconds of mission $M^A_1$ under two different agent controller gains $k_p\in\{4,40\}$.
From $t=0$ s to $t=0.4$ s the residual converges towards zero.
After $t=0.4$ s the residual under $k_p=4$ oscillates between $0$ and $1.47$, while increasing $k_p$ to $40$ leads to a decrease in the residual oscillations by a factor of 10. 
These oscillations may be eliminated by using a different controller in \eqref{eq:X_impl} with knowledge of the Lissajous reference dynamics, e.g. an internal model based controller.

Consequently, for small values of $(\tau_p,\tau_c)$ the results of Theorem \ref{thm:ugasA} are validated. 
Note that (as expected) if large values are chosen for $(\tau_p,\tau_c)$ the results of Theorem are no longer guaranteed, and this is visible in Figure \ref{fig:CLmissions_20} where $q_c$ no longer reaches $\mathcal{A}_c$.

\begin{figure}[b]
    \centering
    \includegraphics[width=0.8\columnwidth]{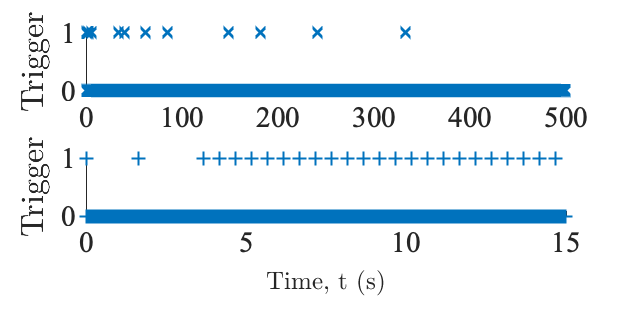}
  \caption{Triggering in Mission $M^C_1$ for the controller sampler for entire trajectory (top), and plant sampler for first 30 seconds (bottom).}
  \label{fig:CLM3timer} 
\end{figure} 

Finally we investigate how the choice of ($\tau_c$,$\tau_p)$ for a mission impacts the frequency of sampling events.
Figure \ref{fig:CLM3timer} depicts when samples are triggered for the plant sampler and controller sampler in Mission $M^C_1$.
As $x_c$ converges to steady state, the controller sampler triggers less often than under sampling with a period of $\min(\tau_c,\tau_p)$ and does not always trigger simultaneously with the plant sampler.
On the other hand, the plant sampler output error periodically exceeds the error threshold after $t=5$ s because $x_p$ tracks a continually-changing Lissajous loop reference. 
As with the residual oscillations discussed above, this can be reduced with the use of a more sophisticated controller incorporating knowledge of the reference trajectory dynamics. 

Together these results highlight how using our proposed sampling scheme with both interfaces can free up more cognitive resources for the commander to apply to secondary tasks.

\section{Conclusions}\label{sec:concl}

In this work we have presented a hybrid systems framework that captures both the interconnected dynamics underlying human-autonomous system interactions and the event-triggered nature of human-on-the-loop systems.
We have also established properties for uniform global asymptotic stability of state solutions under asynchronous event-triggered sampling using the framework.
In practice such a framework may free up cognitive and computational resources that a commander could reallocate to concurrent tasks.
Furthermore, the framework could assist in developing minimal requirements for human-on-the-loop systems with a population of commanders, or in optimizing autonomous system interfaces for a single commander's trust dynamics.

\section*{Acknowledgements}
This paper was written on the lands of the Boonwurrung and Wurundjeri Woiwurrung people.
The authors thank Dr Romain Postoyan, Prof Dragan Nešić, and Dr Elena Vella for discussions and comments.

\bibliographystyle{IEEEtran}
\bibliography{main}

\end{document}